\documentclass{article} % For LaTeX2e
\usepackage{iclr2020_conference,times}

\usepackage{amsmath,amsfonts,bm}

\def\eqref#1{equation~\ref{#1}}
\def\1{\bm{1}}

\DeclareMathAlphabet{\mathsfit}{\encodingdefault}{\sfdefault}{m}{sl}
\SetMathAlphabet{\mathsfit}{bold}{\encodingdefault}{\sfdefault}{bx}{n}

\DeclareMathOperator*{\argmin}{arg\,min}

\usepackage{hyperref}
\usepackage{url}
\usepackage{algorithm}
\usepackage{algorithmic}

\usepackage{graphicx}
\usepackage{caption}
\usepackage{subcaption}

\usepackage{amssymb}
\usepackage{amsmath,bm}
\usepackage{enumitem}
\newtheorem{theorem}{Theorem}

\newtheorem{lemma}{Lemma}

\newtheorem{proposition}{Proposition}

\usepackage[utf8]{inputenc} % usually not needed (loaded by default)
\usepackage[T1]{fontenc}

\usepackage{booktabs}

\usepackage{mathtools}

\title{Predicting with High Correlation Features}

\iclrfinalcopy
\author{Devansh Arpit, Caiming Xiong, Richard Socher \\
Salesforce Research \\
\texttt{darpit@salesforce.com} \\
}

\begin{document}

\maketitle
\begin{abstract}
It has been shown that instead of learning actual object features, deep networks tend to exploit non-robust (spurious) discriminative features that are shared between training and test sets. Therefore, while they achieve state of the art performance on such test sets, they achieve poor generalization on out of distribution (OOD) samples where the IID (independent, identical distribution) assumption breaks and the distribution of non-robust features shifts. In this paper, we consider distribution shift as a shift in the distribution of input features during test time that exhibit low correlation with targets in the training set. Under this definition, we evaluate existing robust feature learning methods and regularization methods and compare them against a baseline designed to specifically capture high correlation features in training set. As a controlled test-bed, we design a colored MNIST (C-MNIST) dataset and find that existing methods trained on this set fail to generalize well on an OOD version this dataset, showing that they overfit the low correlation color features. This is avoided by the baseline method trained on the same C-MNIST data, which is designed to learn high correlation features, and is able to generalize on the test sets of vanilla MNIST, MNIST-M and SVHN datasets. {Our code is available at \url{https://github.com/salesforce/corr_based_prediction}}

\end{abstract}

\section{Introduction}
It is known that deep networks trained on clean training data (without proper regularization) often learn spurious (non-robust) features which are features that can discriminate between classes but do not align with human perception \citep{jo2017measuring,geirhos2018imagenet,tsipras2018robustness,ilyas2019adversarial}. An example of non-robust feature is the presence of desert in camel images, which may correlate well with this object class. More realistically, models can learn to exploit the abundance of input-target correlations present in datasets, not all of which may be invariant under different environments. Interestingly, such classifiers can achieve good performance on test sets which share the same non-robust features. However, due to this exploitation, these classifiers perform poorly under distribution shift \citep{geirhos2018imagenet,hendrycks2019benchmarking} because it violates the IID assumption which is the foundation of existing generalization theory \citep{bartlett2002rademacher,mcallester1999some,mcallester1999pac}.

The research community has approached this problem from different directions. In part of domain adaptation literature (Eg. \cite{ganin2014unsupervised}), the goal is to adapt a model trained on a source domain (often using unlabeled data) so that its performance improves on a target domain that contains the same set of target classes but under a distribution shift. There has also been research on causal discovery \citep{hoyer2009nonlinear,janzing2009telling,lopez2017discovering,kilbertus2018generalization} where the problem is formulated as identifying the causal relation between random variables. This framework may potentially then be used to train a model that only depends on the relevant features. However, it is often hard to discover causal structure in realistic settings. Adversarial training \citep{goodfellow2014explaining,madry2017towards} on the other hand aims to learn models whose predictions are invariant under small perturbations that are humanly imperceptible. Thus adversarial training can be seen as the worst-case distribution shift in the local proximity of the original training distribution.

We consider the situation in which the distribution of input features that have low correlation with labels in training set undergo a shift in their distribution during test time. The intuition behind picking this definition of distribution shift is that dominant correlations are by definition present more universally across samples and should therefore be relatively more representative of the correct label. Under this situation, we study the behavior of existing regularization techniques designed for robust feature learning and avoiding overfitting, and compare it against a baseline that is designed to find input features that correlate strongly with corresponding labels. Our experimental results show that deep network trained using existing regularization methods on a colored version of MNIST dataset (see appendix \ref{sec_cmnist_generation} for samples) are unable to generalize well on a distribution-shifted version of the colored MNIST dataset, while the baseline method generalizes well on this test set along with the test sets of vanilla MNIST, MNIST-M, SVHN .

\section{Baseline Method: Identifying Dominant Correlations}

Here we formulate the objective of the baseline regularization method we use in our experiments which is aimed at learning features that correlate strongly with labels. We then study the behavior of this baseline method. Specifically, let $f(\mathbf{x})$ represent the prediction of our model, then the objective of the baseline is to minimize,
\begin{align}
\label{eq_dominant_corr}
    J(\theta) &= \mathbb{E}[ (f_\theta(\mathbf{x}) -y )^2] +  \frac{\beta}{K}\sum_{k=1}^K \mathbb{E}_{\mathbf{x}\sim \mathcal{D}(\mathbf{x}|y=k)}[(f_\theta(\mathbf{x}) - {\mu}_k)^2] + \lambda \lVert \theta \rVert^2
\end{align}

where ${\mu}_k := \mathbb{E}_{\mathbf{x}\sim \mathcal{D}(\mathbf{x}|y=k)}[ f_\theta(\mathbf{x})]$ and $\mathcal{D}$ denotes the data distribution. For the purpose of analysis on synthetic datasets in this section, we use a linear model $f_{\theta} (\mathbf{x}) := \theta^T \mathbf{x}$. When using a deep network in experiments, we apply the correlation based regularization corresponding to $\beta$ to the first hidden layer of the network. For convolutional networks, a mini-batch has dimensions $(B, C, H, W)$, where we denote B-- batch size, C-- channels, H-- height, W-- width. In this case, we reshape this tensor to take the shape $(B\times H \times W, C)$ and treat each row as a hidden vector $\mathbf{h}$.

% \begin{align}
%     \mathcal{R}_{EP} = \sum_{k=1}^K \mathbb{E}_{\mathbf{x}\sim \mathcal{D}(\mathbf{x}|y=k)}[ \lVert \mathbf{h}(\mathbf{x}) - \mathbf{\mu}_k\rVert^2]
% \end{align}

\subsection{Theoretical Analysis}
\label{sec_theory}

We now theoretically study the behavior of Eq. \ref{eq_dominant_corr} on two synthetic datasets designed to provide insights into the subjective quality of the representation learned.

\subsubsection{Synthetic Dataset A}
\label{sec_synth_dataA}

The baseline regularization should encourage the neural network to pick features that are dominantly present in class samples and able to discriminate between samples from different classes. To formalize the above intuition, we consider the following synthetic data generating process where the data samples $\mathbf{x} \in \mathbb{R}^d$ and labels $y$ are sampled from the following distribution,
\begin{align}
\label{eq_synth_dataA}
    y \sim \{-1,1\} \mspace{40mu} x_i \sim
    \begin{cases}
    \mathcal{N}(y,\sigma^2) & \text{with probability $p_i$} \\
    \mathcal{N}(-y,\sigma^2) & \text{with probability $1-p_i$} \\
    \end{cases}
\end{align}
where $i \in \{1,2, \cdots,d\}$, $y$ is drawn with uniform probability, and $\mathbf{x} = [x_1, x_2, \cdots, x_d]$ is a data sample. Also, all $x_i|y$ are independent of each other. Thus depending on the value of $p_i$, a feature $x_i$ has a small or large amount of information about the label $y$. Specifically, values of $p_i$ close to 0.5 do not tell us anything about the value of $y$ while values close to 0 and 1 can reliably predict its value. Here we make the assumption that features with $p_i$ closer to $0.5$ are non-robust features whose distribution may shift during test time, while features  with $p_i$ closer to $0$ and $1$ are robust ones. Thus we would ideally want a trained model to be insensitive to non-robust features.
The theorem below shows how the model parameters depend on input dimensions for the optimal parameters when training a linear regression model $f_{\theta} (\mathbf{x}) := \theta^T \mathbf{x}$ using the baseline objective.

\begin{theorem}
\label{theorem_dataA}
Let $\theta^{*}$ be the minimizer of $J(\theta)$ in Eq. \ref{eq_dominant_corr} where we have used synthetic dataset A. Then for a large enough $d$, $\theta^* =  \mathbf{M}^{-1} |2\mathbf{p} - \mathbf{1}|$, where $\mathbf{M} := \mathbf{\Sigma} + \lambda \mathbf{I} + \beta(\sigma^2\mathbf{I} + 4\text{diag}(\mathbf{p} \odot (\mathbf{1} - \mathbf{p})))$,

such that $\mathbf{\Sigma}$ is a \textit{positive definite} matrix if\footnote{This assumption is needed due to technicality.} $p_i \not\in \{ 0,0.5,1\}$ for all $i$.
\end{theorem}

As an implication of the above statement, since $\mathbf{M}^{-1}$ is a full rank matrix, aside from the effects due to $\mathbf{\Sigma}$ (which is data dependent and beyond our control), $\theta_i^*$ can in general be non-zero for all input and output correlations. This is especially the case when $\beta=0$ (no regularization). When using a sufficiently large $\beta$, we find that $\theta_i^*$ gets reduced for larger values of $p_i(1-p_i)$, i.e., when $p_i$ is closer to 0.5. Thus the baseline regularization helps suppress dependence of the learned model on non-robust (low correlation) features. 

We also verify that this behavior also holds for deep networks. In this case the regularization is applied to the representation of the first hidden layer of the network. We conduct experiments with both linear and deep models on samples drawn randomly from synthetic dataset A. The input data is in 500 dimensions and we set the value of $p_i$ for each $i$ to be uniformly from $[0,1]$ and fix it henceforth. We then randomly sample 15000 input-target pairs from this dataset with $\sigma^2 = 0.0001$. We train a linear regression model and a 3 hidden layer perceptron (MLP) of width 200 with ReLU activation for 1000 iterations using Adam optimizer with learning rate 0.001 and weight decay 0.00001. 
% Details of the experimental setup can be found in appendix \ref{sec_implementation_details}.

\begin{figure}
\centering
\begin{subfigure}{.5\textwidth}
  \centering
  \includegraphics[width=1\linewidth]{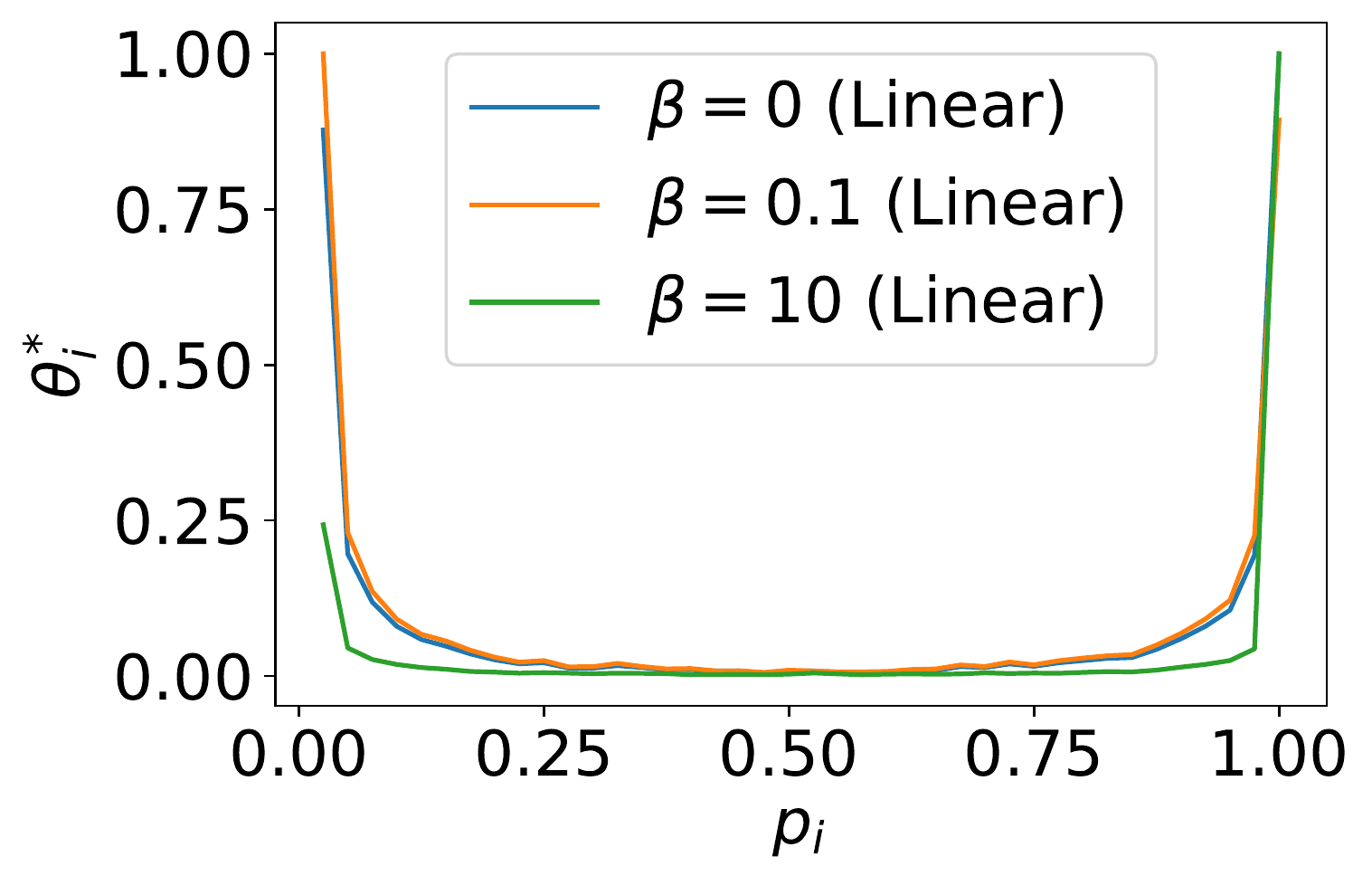}
\end{subfigure}%
\begin{subfigure}{.5\textwidth}
  \centering
  \includegraphics[width=1\linewidth]{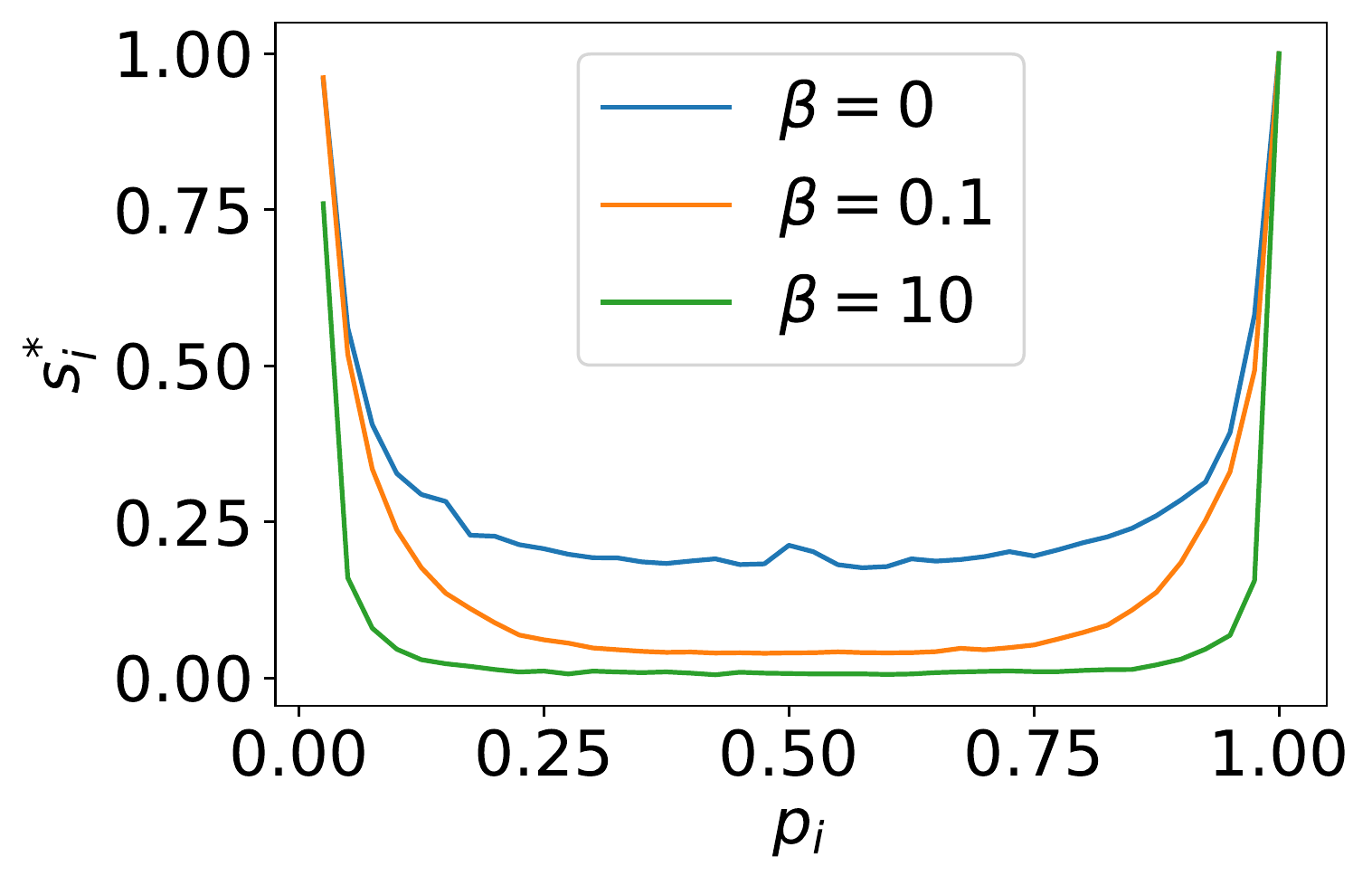}
\end{subfigure}
\caption{Sensitivity $s_i^*$ of output $f_{\theta^*}(\mathbf{x})$ with respect to input dimensions $x_i$ vs. the probability $p_i$ (controlling correlation between input dimension $i$ and target) for synthetic dataset A (Eq. \ref{eq_synth_dataA}). Left plot shows $\theta_i^*$ (same as sensitivity) computed for a trained linear model. Right plot shows sensitivity computed for a trained MLP. The baseline regularization acts as a filter, suppressing the sensitivity of both these models to weak correlation features ($p_i$ close to 0.5).}
\label{fig_dataA_sensitivity}
\end{figure}

In figure \ref{fig_dataA_sensitivity} (left), we plot the parameters $\theta_i^*$ vs. $p_i$ for the linear regression model. Since the same analysis cannot be done for deep networks, we use the perspective that the output-input sensitivity $\mathbf{s}^*$, where $s_i^* := \mathbb{E}_{\mathbf{x}} \left[\bigg| \frac{\partial f_{\theta^*}(\mathbf{x})}{\partial {x}_i} \bigg| \right]$, is equal to $\theta_i^*$ for linear regression. So for deep networks, we plot $s_i^*$ vs. $p_i$ instead as shown in figure \ref{fig_dataA_sensitivity} (right). In both models, we normalize the sensitivity values so that the maximum value is 1 for the ease of comparison across different $\beta$ values. Both for linear and deep models, we find that the sensitivity profile goes to 0 away from $p_i=0$ and 1 when applying the baseline regularization with larger values of coefficients $\beta$; this effect being more dramatic for deep networks. Thus the baseline regularization helps suppress the dependence of model on non-robust (low correlation) features.

\subsubsection{Synthetic Dataset B}
\label{sec_synth_dataB}

We now consider the following binary classification problem where the data samples $\mathbf{x} \in \mathbb{R}^d$ and labels $y$ are sampled from the following distribution,
\begin{align}
    \label{eq_synth_dataB}
    y \sim \{-1,1\} \mspace{40mu} x_i \sim
    \begin{cases}
    \mathcal{N}(y,\sigma^2) & \text{with probability $p_i$} \\
    \mathcal{N}(y,k\sigma^2) & \text{with probability $1-p_i$} \\
    \end{cases}
\end{align}
where $i \in \{1,2, \cdots,d\}$, $y$ is drawn with uniform probability, and $\mathbf{x} = [x_1, x_2, \cdots, x_d]$ is a data sample. Once again, all $x_i|y$ are independent of each other. Thus depending on the value of $p_i$ and $k$, a feature $x_i$ has a small or large variance. We would ideally like the model to avoid dependence on dimensions with high variance because they are non-robust and therefore have a lower correlation with the label. The theorem below shows how the model parameters depend on input dimensions for the optimal parameters when training a linear regression model $f_{\theta} (\mathbf{x}) := \theta^T \mathbf{x}$ using the baseline objective.

\begin{theorem}
\label{theorem_dataB}
Let $\theta^{*}$ be the minimizer of $J(\theta)$ in Eq. \ref{eq_dominant_corr} where we have used synthetic dataset B. Then for a large enough $d$, $ \theta^* =  \mathbf{M}^{-1} \mathbf{1}$, where, $\mathbf{M} := \mathbf{\Sigma} + \lambda \mathbf{I} + \beta\sigma^2 \text{diag}(\mathbf{p} + k(\mathbf{1} - \mathbf{p}))$, such that $\mathbf{\Sigma}$ is a \textit{positive definite} matrix.
\end{theorem}

Once again, we find that $\theta_i^*$ is non-zero for all dimensions of the input. Assume without loss of generality that $k>1$. Then using a sufficiently large $\beta$ would make the value of $\theta_i^*$ approach 0 if $p_i$ is close to 0. In other words, the baseline regularization forces the model to be less sensitive to features with high variance since they are less correlated to label. Thus, such a model's prediction will not be affected significantly under a shift of the distribution of high variance features during test time.

To study the extent of similarity of this behavior between linear regression and deep networks, we once again conduct experiments with both these models on a finite number of samples drawn randomly from synthetic dataset B with $k=10$ and $\sigma^2 = 0.001$. The rest of the details regarding dataset generation and models and optimization are identical to what was used in section \ref{sec_synth_dataA}.

\begin{figure}
\centering
\begin{subfigure}{.35\textwidth}
  \centering
  \includegraphics[width=1.2\linewidth]{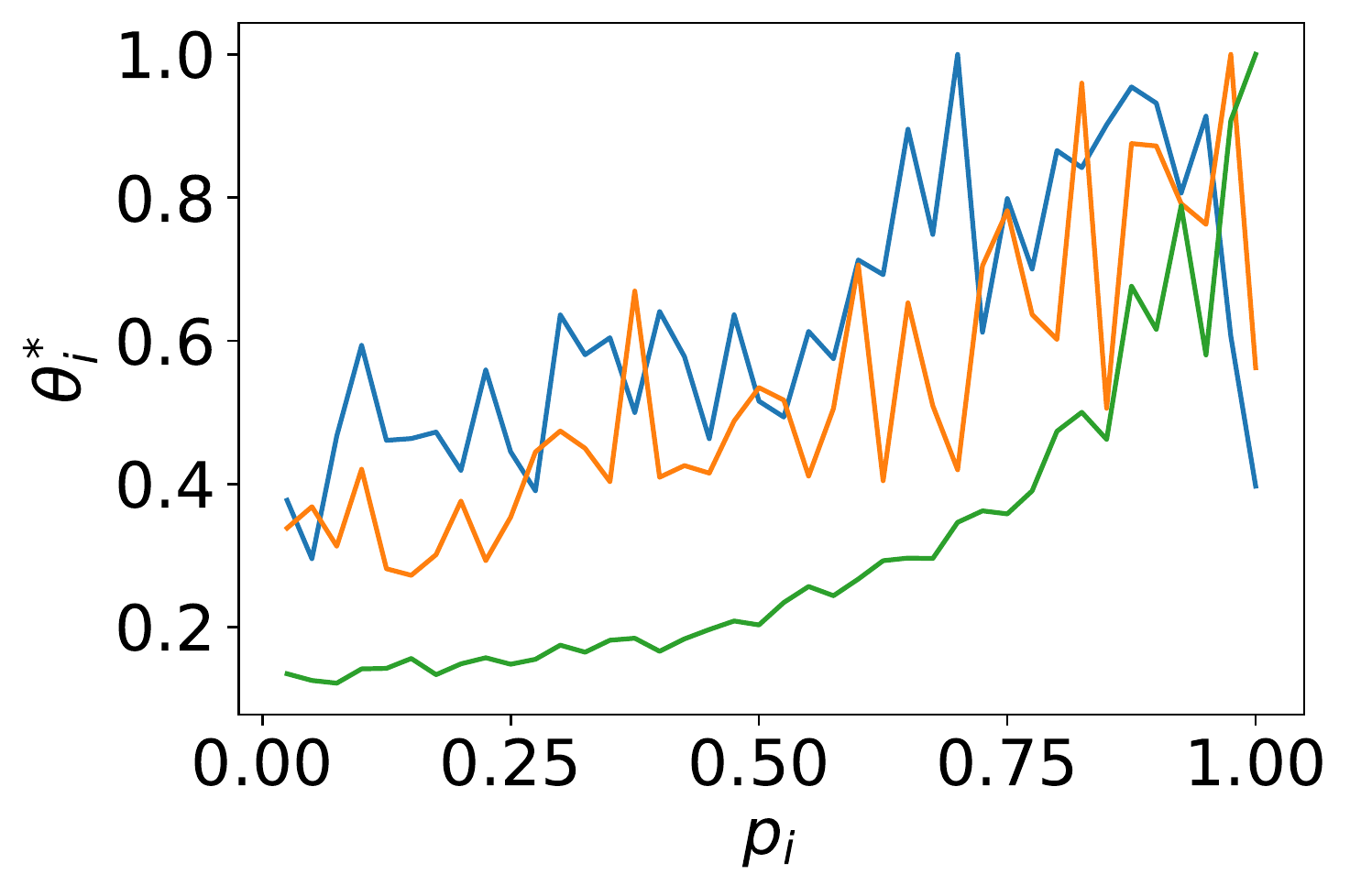}
\end{subfigure}%
\hfill
\begin{subfigure}{.65\textwidth}
  \centering
  \includegraphics[width=0.8\linewidth]{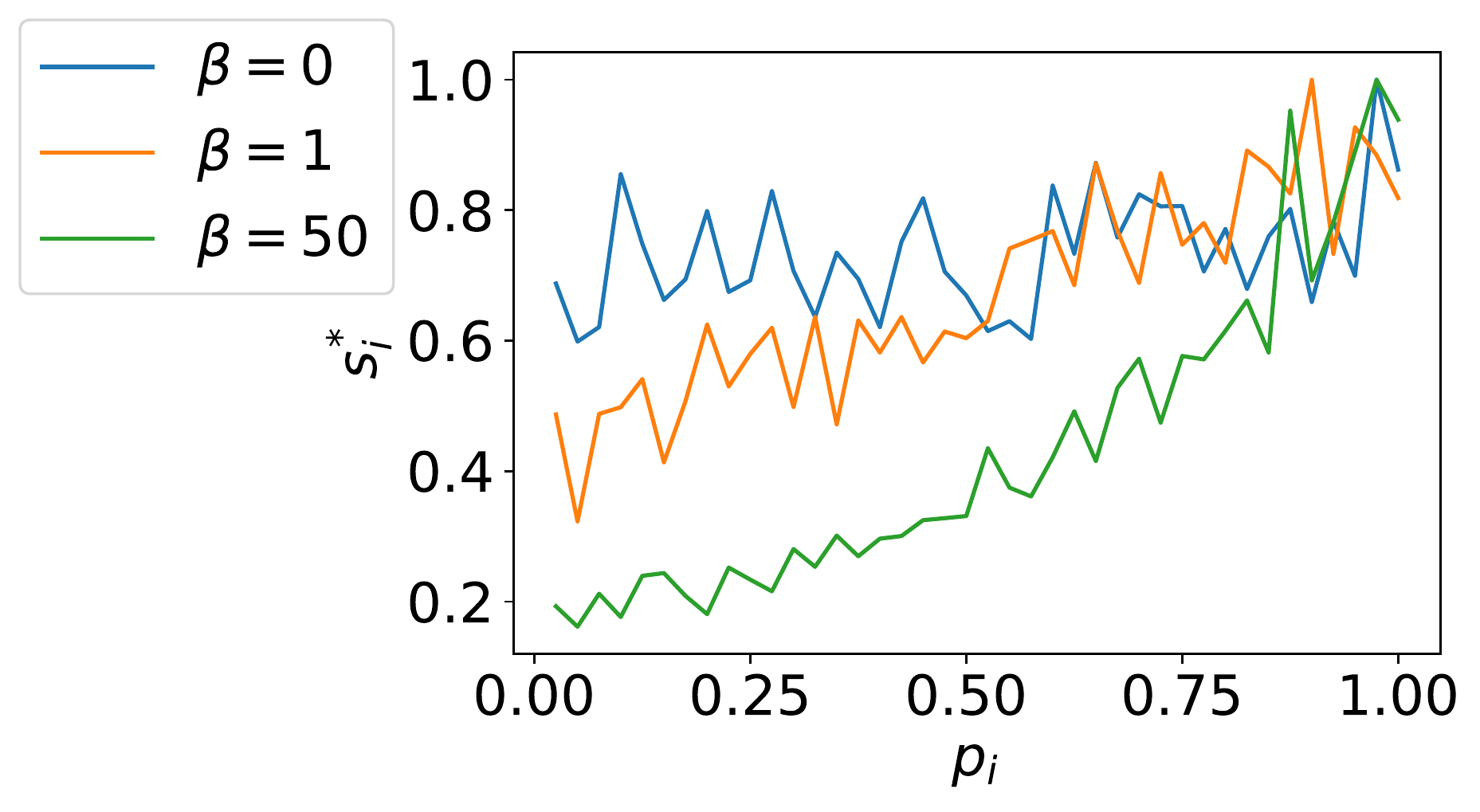}
\end{subfigure}
\caption{Sensitivity $s_i^*$ of output $f_{\theta^*}(\mathbf{x})$ with respect to input dimensions $x_i$ vs. the probability $p_i$ (deciding the choice between feature with variance $\sigma^2$ vs. $10\sigma^2$) for synthetic dataset B (Eq. \ref{eq_synth_dataB}). Left plot shows $\theta_i^*$ (same as sensitivity) computed for a trained linear model. Right plot shows sensitivity computed for a trained MLP. Baseline regularization suppresses the sensitivity of both these models to large variance features ($p_i$ close to 0).}
\label{fig_dataB_sensitivity}
\end{figure}

The sensitivity $s_i^*$ vs. $p_i$ plots are shown in figure \ref{fig_dataB_sensitivity} (left) for linear regression and figure \ref{fig_dataB_sensitivity} (right) for MLP. In the case of linear regression $s_i^* = \theta_i^*$. For both linear regression and MLP, the model's sensitivity to all features are high irrespective of $p_i$ when trained without the baseline regularization ($\beta=0$) and this is especially more so for the MLP. On the other hand, when training with the baseline regularization, we find that a larger $\beta$ forces the models to be less sensitive to input feature dimensions with higher variance (which correspond to $p_i=0$).

\section{Experiments with Data Distribution Shift}
\label{sec_experiments}
% \subsection{Setup}
The experiments below are aimed at investigating the ability of existing regularization methods to generalize when the distribution of low correlation features shift compared to the baseline method. Details not mentioned in the main text can be found in appendix \ref{sec_implementation_details}.

\textbf{Datasets}: We use a colored version of the MNIST dataset (see appendix \ref{sec_cmnist_generation} for dataset samples and details) for experiment 1, and MNIST-M \citep{ganin2016domain}, SVHN \citep{netzer2011reading}, MNIST \citep{lecun-mnisthandwrittendigit-2010} in addition to C-MNIST for experiment 2. All image pixels lie in 0-1 range and are not normalized. The reason for this is that since we are interested in out of distribution (OOD) classification, the normalization constants of training distribution and OOD may be different, in which case data normalized with different statistics cannot be handled by the same network easily.

\textbf{Other Details}: We use ResNet-56 \citep{he2016deep} in all our experiments. We use Adam optimizer \citep{kingma2014adam} with batch size 128 and weight decay 0.0001 for all experiments unless specified otherwise. We do not use batch normalization in any experiment except for the adaptive batch normalization method. Discussion and experiments around batch normalization can be found in appendix \ref{sec_bn_effects}. We do not use any bias parameter in the network because we found it led to less overfitting overall. For all configurations specified for baseline method and existing methods below, the hyper-parameter learning rate was chosen from $\{0.0001, 0.001\}$ unless specified otherwise. For baseline method, the regularization coefficient is chosen from $\{0.1,1,10\}$.

\textbf{Existing methods}:

1. Vanilla maximum likelihood (MLE) training: Since there are no regularization coefficients in this case, we search over batch sizes from $\{ 32,64,128 \}$ for each learning rate value.

2. Variational bottleneck method (VIB, \cite{alemi2016deep}) is a method that minimizes the information bottleneck objective and thus acts as a regularization. The regularization coefficient for VIB is chosen from the set $\{0.01, 0.1, 1, 5 \}$.

3. Clean logit pairing (CLP): Proposed in \cite{kannan2018adversarial}, this method minimizes the $\ell^2$ norm of the difference between the logits of different samples. As shown in proposition \ref{prop_var_ent_eq} (in appendix), minimizing this $\ell^2$ norm is equivalent to minimizing variance of representation in logit space under the assumption that this distribution is Gaussian. In contrast we apply this regularization in the first hidden layer. Due to this similarity, we consider CLP in our experiments. The regularization coefficient for CLP is chosen from $\{ 0.1, 0.5, 1, 10 \}$.

4. Projected gradient descent (PGD) based adversarial training \citep{madry2017towards} has been shown to yield human interpretable features. This makes it a good candidate for investigation. For PGD, $\ell_{\inf}$ perturbation is used with a maximum perturbation $\epsilon$ from the set $\{ 8, 12, 16, 20 \}$ and step size of 2, where all these numbers are divided by 255 since the input is normalized to lie in $[0,1]$ . The number of PGD steps is chosen from the set $\{20, 50 \}$. We randomly choose 12 different configurations out of these combinations.

5. Adversarial logit pairing (ALP, \cite{kannan2018adversarial}) is another approach for adversarial robustness and an alternative to PGD. Since it has the most number of hyper-parameters, we tried a larger number of configurations for this method. Specifically, we use $\ell_{\inf}$ norm with a maximum perturbation $\epsilon$ from the set $\{ 8, 16, 20 \}$ and step size of 2, where all these numbers are divided by 255 since the input is normalized to lie in $[0,1]$ . The number of PGD steps is chosen from the set $\{20, 50 \}$. The regularization coefficient is chosen from $\{0.1,1,10\}$. We randomly choose 15 different configurations out of these combinations.

6. Gaussian Input Noise has been shown to have a similar effect as that from adversarial training \citep{ford2019adversarial} with even better performance in certain cases. We choose Gaussian input noise with standard deviation from the set $\{ 0.05, 0.1, 0.2, 0.3 \}$.

7. Adaptive batch normalization (AdaBN, \cite{li2016revisiting}) has been proposed as a simple way to achieve domain adaptation in which the running statistics of batch normalization are updated with the statistics of the target domain data. Since there are no regularization coefficients in this case, we search over batch sizes from $\{ 32,64,128 \}$ for each learning rate value.

\begin{table}
  \begin{minipage}{0.65\linewidth}
    \centering
    \includegraphics[width=1\linewidth]{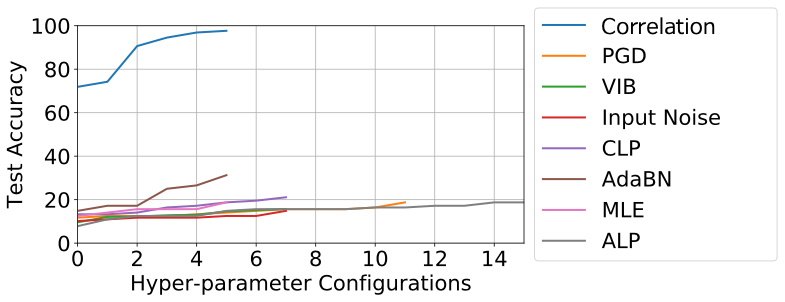}
    \captionof{figure}{Performance on the distribution shifted test set of C-MNIST for various methods trained on C-MNIST training set. See figure \ref{fig_cmnist} in appendix for samples from C-MNIST dataset.}
    \label{fig_cmnist_acc}
  \end{minipage}\hfill
  \begin{minipage}{0.3\linewidth}
    \centering
    \begin{tabular}{lr}
      Dataset          & Accuracy \\
      \midrule
      C-MNIST & 96.88\\
      MNIST     & 93.75 \\
      MNIST-M  & 85.94 \\
      SVHN  & 60.94 \\
      \bottomrule
    \end{tabular}
    \vspace{20pt}
    \caption{Out of distribution performance on test sets using a model trained with baseline method on C-MNIST dataset.}
    \label{table:ood_acc}
  \end{minipage}
\end{table}

\begin{figure*}
\centering
\begin{subfigure}{1\textwidth}
  \centering
  \includegraphics[width=1\linewidth]{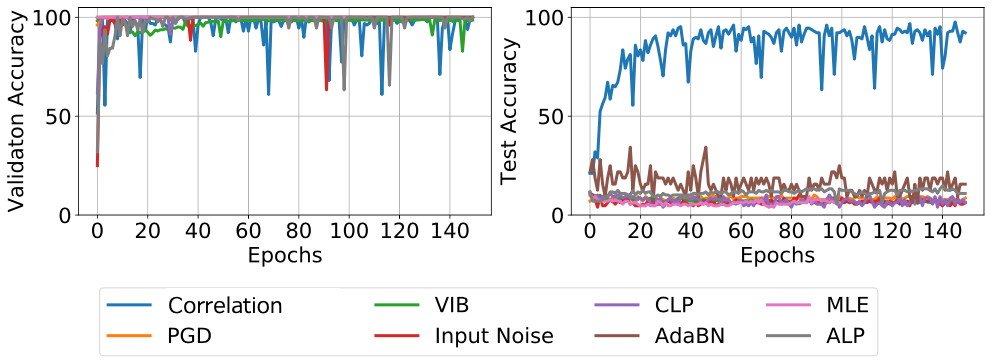}
\end{subfigure}%
\caption{Existing methods severely overfit color features in the C-MNIST training set leading to near $100\%$ accuracy on C-MNIST validation set but close to chance performance on the distribution shifted C-MNIST test set.}
\label{fig_overfitting_color}
\end{figure*}

\textbf{Experiment 1}: In this experiment, we train ResNet-56 on the colored MNIST dataset using the existing methods and baseline regularization, and test the performance of the trained models on the distribution shifted test set of colored MNIST dataset in each case. For each method, we record the best test performance for each hyper-parameter configuration used, and after sorting these numbers across all configurations, we plot them in figure \ref{fig_cmnist_acc}. We find that all the existing methods tend to severely overfit the non-robust color features in C-MNIST leading to poor performance on the distribution shifted test set of C-MNIST except the baseline method (named correlation). 

To further confirm this claim, we plot the validation and test accuracy vs. epoch for all methods for one of the hyper-parameter configurations. The validation set of C-MNIST is a held out set that follows the same distribution as the training set but the two are mutually exclusive. On the other hand, the test set of C-MNIST has different foreground and background colors in all images that are sampled independently of the training set as explained in section \ref{sec_cmnist_generation} and shown in figure \ref{fig_cmnist}. The hyper-parameter configuration chosen (from among the ones used in experiment 1 in section \ref{sec_experiments}) for each method is based on the condition that training accuracy converged to $100\%$ at the end of the training process. This ensures a fair comparison of validation and test accuracy between different methods. The plots can be seen in figure \ref{fig_overfitting_color}. Clearly, existing methods achieve near $100\%$ accuracy on C-MNIST validation set but close to chance performance on the distribution shifted C-MNIST test set, showing that these methods have overfitted the color features. Correlation based learning is able to avoid this dependence.

\textbf{Experiments 2}: In this experiment, we hand-pick the model trained with the baseline regularization on C-MNIST in experiment 1 above, such that it simultaneously performs well on SVHN, MNIST-M and MNIST datasets. We used the C-MNIST test set for early stopping. These performances are shown in table \ref{table:ood_acc}. These experiments show that the dominant correlation based features learned by the baseline model on C-MNIST dataset capture shape features which allow the model to generalize reasonably well on the other digit datasets with drastic distribution shifts.

\section{Related Work}
\label{sec_related_work}

\textbf{Invariant Risk Minimization}: 
The goal of IRM \citep{arjovsky2019invariant} is to achieve out of distribution generalization under the formalism that certain features have a stable correlation with target across all possible environments. In other words, if there are multiple features that correlate with label, then IRM aims to learn the feature which has the same degree of correlation with label irrespective of the environment, while ignoring other features. IRM achieves this goal by learning representations such that there exists a predictor (Eg. a linear classifier) that is simultaneously optimal for representations across all environments. We instead explore the idea of extracting features that have high correlation with labels.

\textbf{Adversarial Training}: 
There is an abundance of literature around robust optimization \citep{wald1945statistical,BEN_RO} and adversarial training \citep{goodfellow2014explaining,madry2017towards} which study robustness of models to small perturbations around input samples and are often studied using first order methods. Such perturbations can be seen as the worst case distribution shift in the local proximity of the original training distribution. Further, \cite{tsipras2018robustness} discusses that the representations learned by adversarially trained deep network are more human interpretable. These factors make it a good candidate for investigating its behavior under distribution shift.

\textbf{Domain Adaptation}: Domain adaptation \citep{wang2018deep,patel2014visual} addresses the problem of distribution shift between source and target domain, and has attracted considerable attention in computer vision, NLP and speech communities \citep{kulis2011you,blitzer2007biographies,hosseini2018multi}. Some of these methods address this issue by aligning the two distributions \citep{jiang2007instance,bruzzone2009domain}, while others by making use of adversarial training \citep{ganin2014unsupervised, ganin2016domain} and auxilliary losses \citep{ghifary2015domain, he2016dual}. A common characteristic of all these methods is that they require labeled/unlabeled target domain data during the training process.

\section{Conclusion}
\label{sec_discussion}
We explored the idea of using input feature with high correlation with labels for making prediction such that the distribution of low correlation features shifted during test time. We found that existing methods that are aimed at learning robust representation in the adversarial sense or in the general sense of reducing overfitting are unable to handle such distribution shifts. On the other hand, our regularization specifically designed for this task performed well under this distribution shift during test time.

\section*{Acknowledgments}
We would like to thank Bernhard Geiger, Stanisław Jastrzębski, Ehsan Hosseini-Asl, Lav Varshney and Tong Niu for discussions and feedback.

\bibliography{ref}
\bibliographystyle{iclr2020_conference}

\newpage
\setcounter{page}{1}
\appendix
\onecolumn
\section*{Appendix} 

\section{Datasets}
\label{sec_cmnist_generation}

\begin{figure}
\centering
\begin{subfigure}{.5\textwidth}
  \centering
  \includegraphics[width=0.6\linewidth]{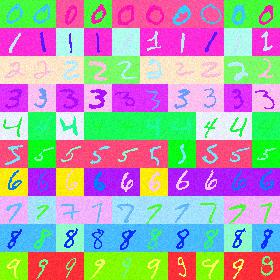}
\end{subfigure}%
\hfill
\begin{subfigure}{.5\textwidth}
  \centering
  \includegraphics[width=0.6\linewidth]{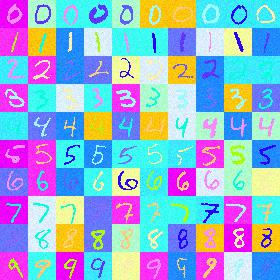}
\end{subfigure}
\caption{Color MNIST training set (left) and out of distribution test set (right). Each class in training set has two background colors and two foreground colors that are unique only to that class. Each test image has a foreground and background color that is randomly picked out of 10 colors that are chosen independently of the training set.}
\label{fig_cmnist}
\end{figure}

\begin{figure}
\centering
\begin{subfigure}{1\textwidth}
  \centering
  \includegraphics[width=0.5\linewidth]{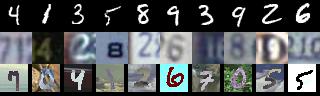}
\end{subfigure}%
\caption{Random samples from MNIST (top), SVHN (middle) and MNIST-M (bottom) datasets are shown to get a visual sense of the hardness of the out of distribution task.}
\label{fig_all_dataset}
\end{figure}

\textbf{Colored MNIST Dataset}: Randomly drawn training and test samples from the C-MNIST dataset generated as described above are shown in figure \ref{fig_cmnist}. The colored MNIST dataset (C-MNIST), which is used in experiments 1 and 2, uses all the 60,000 training image in the MNIST dataset to generate the C-MNIST dataset, where we randomly do a 0.9-0.1 split to get the training set and validation set respectively. Similarly, we use all the 10,000 test images in MNIST to generate the C-MNIST test set. We vary both the foreground and background colors to generate our C-MNIST dataset. The reason why we vary colors of both foreground and background is that we want the trained model to avoid overfitting any color bias. A single color in foreground or background would constitute a low variance feature, which as we study in section \ref{sec_synth_dataB}, leads models to prioritize learning it for both training with vanilla MLE as well as MLE with baseline method.

The C-MNIST training/validation set is generated from its MNIST counter-part as follows:

1. For each class, randomly assign two colors (RGB value) for foreground and two colors (RGB value) for background.

2. Binarize each image (pixels in 0-255 range) around threshold 150 so that pixel values are either 0 or 1 and replicate the channel in each image to have a three channel image.

3. For each image in a class, randomly pick one of the two foreground colors assigned to that class and replace all foreground pixel with that color. Similarly replace background pixels for all images.

4. Add zero mean Gaussian noise with a small standard deviation (0.04 used in our experiments) to all images.

To generate the test set, in step one, we randomly assign a foreground and background color to each image irrespective of the class, and the colors for validation set are chosen independently of the training set.

\textbf{Other Datasets}: During experiments with MNIST, the single channel in each image was replicated to form a three channel image. For SVHN, we resized the image to have $28\times 28$ hieght and width.

\section{Experimental Details}
\label{sec_implementation_details}

\textbf{Variational bottleneck method} (VIB): As an implementation detail of VIB, we flatten the output of the last convolution layer of ResNet-56 and separately pass it through two linear layer of width 256, one that outputs a vector that we regard as mean $\mu$, and the other that we regard as log-variance $\nu$ (similar to how it is done in variational auto-encoders \citep{kingma2013auto}). We then combine their outputs as $\mathbf{o} = \mathbf{\mu} + exp(0.5\mathbf{\nu})\odot \mathbf{\epsilon}$ where $\mathbf{\epsilon}$ is sampled from a standard Gaussian distribution of the same dimension as $\mathbf{\nu}$. This output is then passed through a linear layer that transforms it into a vector of dimension same as the number of classes. These implementation details are similar to those in the original VIB paper.

Further, we gradually ramp up the regularization coefficient $\beta$ from 0.0001 to its final value by doubling the current value at the end of every epoch until the final value is reached. This is a popular prctice when minimizing the KL divergence term between posterior and prior which helps optimization.

\section{Effects of Batch Normalization on Baseline Method}
\label{sec_bn_effects}

\begin{figure}
\centering
\begin{subfigure}{1\textwidth}
  \centering
  \includegraphics[width=1\linewidth]{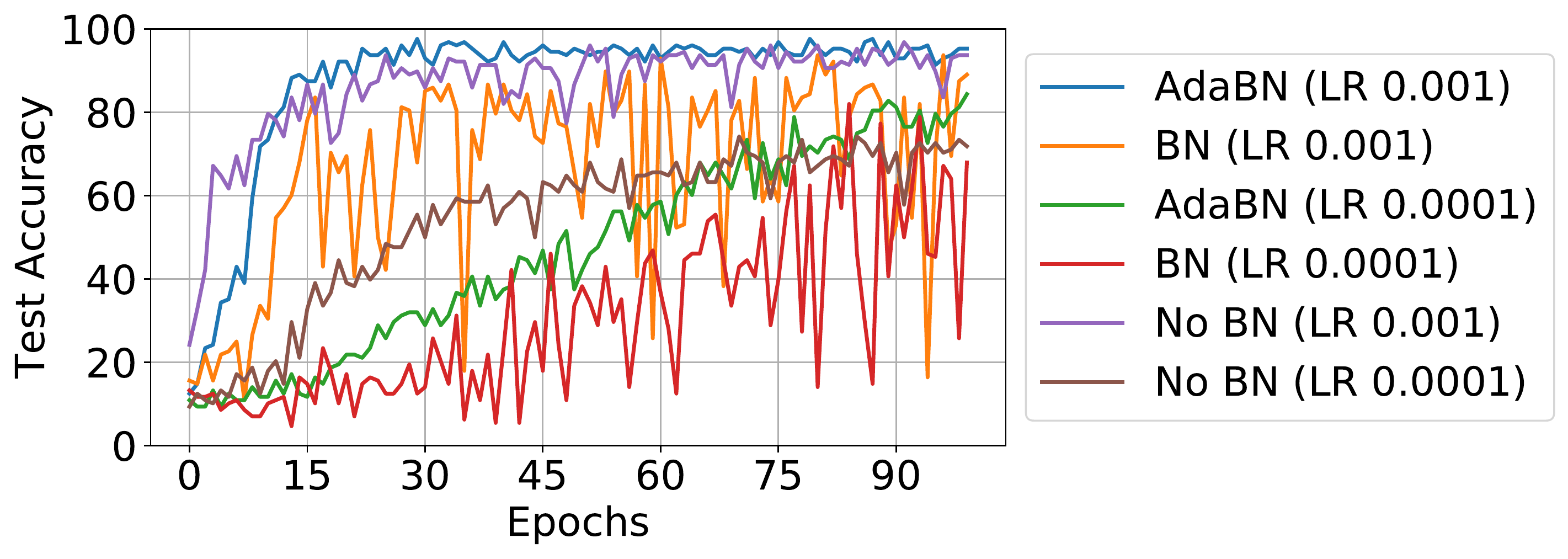}
\end{subfigure}%
\caption{Batch Normalization is unstable without AdaBN when using proposed regularization (experiment on C-MNIST dataset).}
\label{fig_bn_effects}
\end{figure}

In this section we study the effect of batch normalization (BN, \cite{ioffe2015batch}) when using it in conjunction with baseline method. We train a ResNet-56 on C-MNIST dataset identical to the settings in experiment 1 in main text, with the exception that we use BN. While doing so, we record the accuracy of the model on the C-MNIST test set at every epoch. In addition, we also run experiments where under the same training settings, during evaluation at each epoch, we use AdaBN \citep{li2016revisiting}. These values are plotted in figure \ref{fig_bn_effects}. We have also plotted runs using baseline method without any batch normalization for reference. In all cases, we use a learning rates (LR) of 0.001 and 0.0001. The plots show that without AdaBN, test accuracy is very unstable on the distribution shifted test set. This seems to be a side-effect of using BN which can be fixed by adapting the running statistics of BN (used during evaluation) with the test set statistics. However, this requires domain knowledge (i.e., samples) of the test set, which is not preferable for our goal.

\section{Proofs}

\begin{lemma}
\label{lemma_rvstats_dataA}
\begin{align}
    \mathbb{E}[x_i | y=1] = -\mathbb{E}[x_i | y=-1] = 2p_i-1\\
    \mathbb{E}[x_i^2|y=1] = \mathbb{E}[x_i^2|y=-1] = 1 + \sigma^2
\end{align}
\textbf{Proof}: Given the distribution of $\mathbf{x}$, we can write each element ${x}_i$ as,
\begin{align}
    x_i = b_i n_i + (1-b_i)\bar{n}_i
\end{align}
where $b_i$ is sampled from the Bernoilli distribution with probability $p_i$, $n_i \sim \mathcal{N}(y, \sigma^2)$, and $\bar{n}_i \sim \mathcal{N}(-y, \sigma^2)$. Thus,
\begin{align}
    \mathbb{E}[x_i | y=1] &= p_i - (1-p_i) = 2p_i-1\\
    \mathbb{E}[x_i | y=-1] &= -p_i + (1-p_i) = 1-2p_i
\end{align}
Next,
\begin{align}
    \mathbb{E}[x_i^2|y=1] &= \mathbb{E}[b_i^2 n_i^2 + (1-b_i)^2\bar{n}_i^2 + 2b_i(1-b_i)n_i\bar{n}_i|y=1]
\end{align}
We know from the properties of Bernoulli and Gaussian distribution that,
\begin{align}
    \mathbb{E}[b_i^2] &= var(b_i) + \mathbb{E}[b_i]^2 = p_i(1-p_i) + p_i^2 = p_i\\
    \mathbb{E}[n_i^2|y=1] &= var(n_i|y=1) + \mathbb{E}[n_i|y=1]^2 = \sigma^2 + 1
\end{align}
We similarly get, 
\begin{align}
    \mathbb{E}[n_i^2|y=-1] &= \sigma^2 + 1\\
    \mathbb{E}[\bar{n}_i^2|y=1] &= \sigma^2 + 1\\
    \mathbb{E}[\bar{n}_i^2|y=-1] &= \sigma^2 + 1
\end{align}
Therefore, using the independence property between random variables,
\begin{align}
    \mathbb{E}[x_i^2|y=1] &= \mathbb{E}[b_i^2]\mathbb{E}[n_i^2|y=1] + (1+\mathbb{E}[b_i^2] - 2\mathbb{E}[b_i])\mathbb{E}[\bar{n}_i^2|y=1]\nonumber\\ 
    &+ 2(\mathbb{E}[b_i] - \mathbb{E}[b_i]^2)\mathbb{E}[\bar{n}_i|y=1]\mathbb{E}[n_i|y=1]\\
    &= 1 + \sigma^2
\end{align}
We similarly have,
\begin{align}
    \mathbb{E}[x_i^2|y=-1] = 1+\sigma^2
\end{align}
\end{lemma}

\begin{theorem}
Let $\theta^{*}$ be the minimizer of $J(\theta)$ in Eq. \ref{eq_dominant_corr} where we have used synthetic dataset A. Then for a large enough $d$, $\theta^*$ is given by,
\begin{align}
      \theta^* =  \mathbf{M}^{-1} |2\mathbf{p} - \mathbf{1}|
\end{align}
where,
\begin{align}
    \mathbf{M} := \mathbf{\Sigma} + \lambda \mathbf{I} + \beta(\sigma^2\mathbf{I} + 4\text{diag}(\mathbf{p} \odot (\mathbf{1} - \mathbf{p})))
\end{align}
such that $\mathbf{\Sigma}$ is a \textit{positive definite} matrix if\footnote{This assumption is needed due to technicality.} $p_i \not\in \{ 0,0.5,1\}$ for all $i$.

\textbf{Proof}: We note that,
\begin{align}
    f_\theta(\mathbf{x})|(y=1) &= \sum_{i=1}^d \theta_i x_i|y=1\\
    &= \sum_{i=1}^d \theta_i( b_i (n_i |y=1) + (1-b_i)(\bar{n}_i|y=1))
\end{align}
For a large enough dimensionality $d$ of $\mathbf{x}$, central limit theorem (CLT) applies to $f_\theta(\mathbf{x})|(y=1)$ and it converges to a Gaussian distribution. Let $s_1^2$ denote the variance of $f_\theta(\mathbf{x})|y=1$. A similar argument applies to $f_\theta(\mathbf{x})|y=-1$, in which case we define $s_{-1}^2$ to be its variance. Thus,
\begin{align}
    s_1^2 &= var(\sum_{i=1}^d \theta_i x_i | y=1)\\
    &= \sum_{i=1}^d var(\theta_i x_i | y=1)\\
    &= \sum_{i=1}^d \theta_i^2 var(x_i | y=1)
\end{align}
where the second equality holds because $x_i$'s are independent of one another. Thus using lemma \ref{lemma_rvstats_dataA},
\begin{align}
    s_1^2 &= \sum_i \theta_i^2 (1 + \sigma^2 - (2p_i-1)^2)\\
    &= \sum_i \theta_i^2 ( \sigma^2 + 4p_i(1-p_i))
\end{align}
We similarly get,
\begin{align}
    s_{-1}^2 &= \sum_i \theta_i^2 ( \sigma^2 + 4p_i(1-p_i))
\end{align}
Since $s_1^2$ and $s_{-1}^2$ are equal, we denote $s^2 = s_1^2 = s_{-1}^2$.
Therefore, our objective becomes,
\begin{align}
    &\argmin_{\theta} \mathbb{E}[ (f_\theta(\mathbf{x}) -y )^2] + \lambda \lVert \theta \rVert^2 + \beta s^2\\
    &= \argmin_{\theta} \theta^T\mathbb{E}[\mathbf{xx}^T]\theta - 2\theta^T \mathbb{E}[\mathbf{x}y] + \lambda \lVert \theta \rVert^2 + \beta \sum_{i=1}^d \theta_i^2 ( \sigma^2 + 4p_i(1-p_i))
\end{align}
Define $\mathbf{M}$ as,
\begin{align}
    \mathbf{M} := \mathbf{\Sigma} + \lambda \mathbf{I} + \beta(\sigma^2\mathbf{I} + 4\text{diag}(\mathbf{p} \odot (\mathbf{1} - \mathbf{p})))
\end{align}
where $\mathbf{\Sigma}:= \mathbb{E}[\mathbf{xx}^T]$, we can re-write our objective as,
\begin{align}
    & \argmin_{\theta} \theta^T \mathbf{M} \theta - 2\theta^T \mathbb{E}[\mathbf{x}y]
\end{align}
whose solution is given by,
\begin{align}
\label{eq_theta_sol_th1}
      \theta^* =  \mathbf{M}^{-1} \mathbb{E}[\mathbf{x}y]
\end{align}

Using lemma \ref{lemma_rvstats_dataA}, we get,
\begin{align}
    \mathbb{E}[x_iy] &= \mathbb{E}[x_i|y=1]\Pr(y=1) - \mathbb{E}[x_i|y=-1]\Pr(y=-1)\\
    &= 0.5(\mathbb{E}[x_i|y=1] - \mathbb{E}[x_i|y=-1])\\
    &= 0.5(4p_i-2) = 2p_i-1
\end{align}
Plugging this value in Eq. \ref{eq_theta_sol_th1} yields $\theta^*$.

Now we prove that $\mathbf{\Sigma}$ is full rank. Note that it is positive semi-definite since it is a scatter matrix. Next, due to conditional independence, for $i\neq j$,
\begin{align}
    \mathbb{E}[x_i x_j]  &=  \mathbb{E}[x_i x_j|y=1] \Pr(y=1) +  \mathbb{E}[x_i x_j|y=-1] \Pr(y=-1) \\
    &= \mathbb{E}[x_i|y=1] \mathbb{E}[x_j|y=1] \Pr(y=1) +  \mathbb{E}[x_i|y=-1] \mathbb{E}[x_j|y=-1] \Pr(y=-1) 
\end{align}
and for $i= j$,
\begin{align}
    \mathbb{E}[x_i^2]  &=  \mathbb{E}[x_i^2|y=1] \Pr(y=1) +  \mathbb{E}[x_i^2|y=-1] \Pr(y=-1)
\end{align}
Using lemma \ref{lemma_rvstats_dataA}, we get,
\begin{align}
\label{eq_SigmaA}
\Sigma_{ij} = \begin{cases}
    1+\sigma^2 & \text{if $i=j$} \\
     (1-2p_i)(1-2p_j) & \text{otherwise}
    \end{cases}
\end{align}
To prove that $\mathbf{\Sigma}$ is positive definite (and hence full rank), we need to prove that no two columns are parallel. To show this, consider any two indices $i$ and $j$ such that $i\neq j$. We show that there exists no $\alpha \neq 0$ such that the columns $\Sigma_i = \alpha \Sigma_j$. We prove this by contradiction. Suppose $\Sigma_{ii} = \alpha \Sigma_{ji}$, then $\alpha \Sigma_{jj} = \frac{\Sigma_{ii} \Sigma_{jj}}{\Sigma_{ji}}$. Substituting values from Eq. \ref{eq_SigmaA}, 
\begin{align}
    \alpha \Sigma_{jj} = \frac{(1+\sigma^2)^2}{(1-2p_i)(1-2p_j)}
\end{align}
Thus $\alpha \Sigma_{jj}>1$. However, $\Sigma_{ij} = (1-2p_i)(1-2p_j) <1$. Thus there is no non-zero $\alpha$ for which $\Sigma_i = \alpha \Sigma_j$. Hence $\mathbf{\Sigma}$ must be full rank and hence positive definite. Thus we have proved the claim. $\square$
\end{theorem}

%% invariance type 2

\begin{lemma}
\label{lemma_rvstats_dataB}
\begin{align}
    \mathbb{E}[x_i | y=1] = -\mathbb{E}[x_i | y=-1] =1\\
    \mathbb{E}[x_i^2|y=1] = \mathbb{E}[x_i^2|y=-1] = 1 + \sigma^2 (p_i + k_i(1-p_i))
\end{align}
\textbf{Proof}: Given the distribution of $\mathbf{x}$, we can write each element ${x}_i$ as,
\begin{align}
    x_i = b_i n_i + (1-b_i)\bar{n}_i
\end{align}
where $b_i$ is sampled from the Bernoilli distribution with probability $p_i$, $n_i \sim \mathcal{N}(y, \sigma^2)$, and $\bar{n}_i \sim \mathcal{N}(y, k_i\sigma^2)$. Thus,
\begin{align}
    \mathbb{E}[x_i | y=1] &= p_i + (1-p_i) = 1\\
    \mathbb{E}[x_i | y=-1] &= -p_i - (1-p_i) = -1
\end{align}
Next,
\begin{align}
    \mathbb{E}[x_i^2|y=1] &= \mathbb{E}[b_i^2 n_i^2 + (1-b_i)^2\bar{n}_i^2 + 2b_i(1-b_i)n_i\bar{n}_i|y=1]
\end{align}
We know from the properties of Bernoulli and Gaussian distribution that,
\begin{align}
    \mathbb{E}[b_i^2] &= var(b_i) + \mathbb{E}[b_i]^2 = p_i(1-p_i) + p_i^2 = p_i\\
    \mathbb{E}[n_i^2|y=1] &= var(n_i|y=1) + \mathbb{E}[n_i|y=1]^2 = \sigma^2 + 1
\end{align}
We similarly get, 
\begin{align}
    \mathbb{E}[n_i^2|y=-1] &= \sigma^2 + 1\\
    \mathbb{E}[\bar{n}_i^2|y=1] &= k\sigma^2 + 1\\
    \mathbb{E}[\bar{n}_i^2|y=-1] &= k\sigma^2 + 1
\end{align}
Therefore, using the independence property between random variables,
\begin{align}
    \mathbb{E}[x_i^2|y=1] &= \mathbb{E}[b_i^2]\mathbb{E}[n_i^2|y=1] + (1+\mathbb{E}[b_i^2] - 2\mathbb{E}[b_i])\mathbb{E}[\bar{n}_i^2|y=1]\nonumber\\ 
    &+ 2(\mathbb{E}[b_i] - \mathbb{E}[b_i]^2)\mathbb{E}[\bar{n}_i|y=1]\mathbb{E}[n_i|y=1]\\
    &= p_i(1 + \sigma^2) + (1-p_i)(1+k\sigma^2)
\end{align}
We similarly have,
\begin{align}
    \mathbb{E}[x_i^2|y=-1] = p_i(1 + \sigma^2) + (1-p_i)(1+k\sigma^2)
\end{align}
Rearranging these terms yields the claim. $\square$
\end{lemma}

\begin{theorem}
Let $\theta^{*}$ be the minimizer of $J(\theta)$ in Eq. \ref{eq_dominant_corr} where we have used synthetic dataset B. Then for a large enough $d$, $\theta^*$ is given by,
\begin{align}
      \theta^* =  \mathbf{M}^{-1} \mathbf{1}
\end{align}
where,
\begin{align}
    \mathbf{M} := \mathbf{\Sigma} + \lambda \mathbf{I} + \beta\sigma^2 \text{diag}(\mathbf{p} + k(\mathbf{1} - \mathbf{p}))
\end{align}
such that $\mathbf{\Sigma}$ is a \textit{positive definite} matrix.

\textbf{Proof}: Similar to theorem \ref{theorem_dataA} we have that,
\begin{align}
    f_\theta(\mathbf{x})|(y=1) &= \sum_{i=1}^d \theta_i b_i (n_i |y=1) + (1-b_i)(\bar{n}_i|y=1)
\end{align}
For a large enough dimensionality $d$ of $\mathbf{x}$, central limit theorem (CLT) applies to $f_\theta(\mathbf{x})|(y=1)$ and it converges to a Gaussian distribution. Let $s_1^2$ denote the variance of $f_\theta(\mathbf{x})|y=1$. A similar argument applies to $f_\theta(\mathbf{x})|y=-1$, in which case we define $s_{-1}^2$ to be its variance. Thus,
\begin{align}
    s_1^2 &= \sum_{i=1}^d \theta_i^2 var(x_i | y=1)
\end{align}
Thus using lemma \ref{lemma_rvstats_dataB},
\begin{align}
    s_1^2 &= \sum_i \theta_i^2 (\sigma^2 (p_i + k_i(1-p_i)))
\end{align}
We similarly get,
\begin{align}
    s_{-1}^2 &= \sum_i \theta_i^2 (\sigma^2 (p_i + k_i(1-p_i)))
\end{align}
Since $s_1^2$ and $s_{-1}^2$ are equal, we denote $s^2 = s_1^2 = s_{-1}^2$.
Therefore, our objective becomes,
\begin{align}
    & \argmin_{\theta} \mathbb{E}[ (f_\theta(\mathbf{x}) -y )^2] + \lambda \lVert \theta \rVert^2 + \beta s^2\\
    &= \argmin_{\theta} \theta^T\mathbb{E}[\mathbf{xx}^T]\theta - 2\theta^T \mathbb{E}[\mathbf{x}y] + \lambda \lVert \theta \rVert^2 + \beta \sigma^2 \sum_i \theta_i^2  (p_i + k_i(1-p_i))
\end{align}
Define $\mathbf{M}$ as,
\begin{align}
    \mathbf{M} := \mathbf{\Sigma} + \lambda \mathbf{I} + \beta\sigma^2 \text{diag}(\mathbf{p} + k(\mathbf{1} - \mathbf{p}))
\end{align}
where $\mathbf{\Sigma}:= \mathbb{E}[\mathbf{xx}^T]$, we can re-write our objective as,
\begin{align}
    & \argmin_{\theta} \theta^T \mathbf{M} \theta - 2\theta^T \mathbb{E}[\mathbf{x}y]
\end{align}
whose solution is given by,
\begin{align}
\label{eq_theta_sol_th2}
      \theta^* =  \mathbf{M}^{-1} \mathbb{E}[\mathbf{x}y]
\end{align}
Using lemma \ref{lemma_rvstats_dataB}, we get,
\begin{align}
    \mathbb{E}[x_iy] &= \mathbb{E}[x_i|y=1]\Pr(y=1) - \mathbb{E}[x_i|y=-1]\Pr(y=-1)\\
    &= 0.5(\mathbb{E}[x_i|y=1] - \mathbb{E}[x_i|y=-1])\\
    &= 1
\end{align}
Plugging this value in Eq. \ref{eq_theta_sol_th2} yields $\theta^*$.

Now we prove that $\mathbf{\Sigma}$ is full rank. Note that it is positive semi-definite since it is a scatter matrix. Next, due to conditional independence, for $i\neq j$,
\begin{align}
    \mathbb{E}[x_i x_j]  &=  \mathbb{E}[x_i x_j|y=1] \Pr(y=1) +  \mathbb{E}[x_i x_j|y=-1] \Pr(y=-1) \\
    &= \mathbb{E}[x_i|y=1] \mathbb{E}[x_j|y=1] \Pr(y=1) +  \mathbb{E}[x_i|y=-1] \mathbb{E}[x_j|y=-1] \Pr(y=-1) 
\end{align}
and for $i= j$,
\begin{align}
    \mathbb{E}[x_i^2]  &=  \mathbb{E}[x_i^2|y=1] \Pr(y=1) +  \mathbb{E}[x_i^2|y=-1] \Pr(y=-1)
\end{align}
Using lemma \ref{lemma_rvstats_dataB}, we get,
\begin{align}
\label{eq_SigmaB}
\Sigma_{ij} = \begin{cases}
    1 + \sigma^2 (p_i + k_i(1-p_i)) & \text{if $i=j$} \\
     1 & \text{otherwise}
    \end{cases}
\end{align}
To prove that $\mathbf{\Sigma}$ is positive definite (and hence full rank), we need to prove that no two columns are parallel. To show this, consider any two indices $i$ and $j$ such that $i\neq j$. We show that there exists no $\alpha \neq 0$ such that the columns $\Sigma_i = \alpha \Sigma_j$. We prove this by contradiction. Suppose $\Sigma_{ii} = \alpha \Sigma_{ji}$, then $\alpha \Sigma_{jj} = \frac{\Sigma_{ii} \Sigma_{jj}}{\Sigma_{ji}}$. Substituting values from Eq. \ref{eq_SigmaB}, 
\begin{align}
    \alpha \Sigma_{jj} = {(1 + \sigma^2 (p_i + k(1-p_i))) (1 + \sigma^2 (p_j + k(1-p_j)))}
\end{align}
Thus $\alpha \Sigma_{jj}>1$. However, $\Sigma_{ij} =1$. Thus there is no non-zero $\alpha$ for which $\Sigma_i = \alpha \Sigma_j$. Hence $\mathbf{\Sigma}$ must be full rank and hence positive definite. Thus we have proved the claim. $\square$
\end{theorem}

\begin{proposition}
\label{prop_var_ent_eq}
If $\Pr(f_{\theta}(\mathbf{X}))$ follows a Gaussian distribution, then,
\begin{align}
    var(f_{\theta}(\mathbf{X})) =   0.5 \mathbb{E}_{\mathbf{X}_1, \mathbf{X}_2 \sim \mathcal{D}(\mathbf{X})}[(f_{\theta}(\mathbf{X}_1) - f_{\theta}(\mathbf{X}_2))^2]
\end{align}
where $\mathbf{X}_1$ and $\mathbf{X}_2$ are IID samples from the data distribution $\mathcal{D}(\mathbf{X})$.

\textbf{Proof}: 
Denote $\mathbf{\mu}$ and $\sigma^2$ as the mean and variance of the Gaussian distribution $\Pr(f_{\theta}(\mathbf{X}))$ respectively. Since $\mathbf{X}_1$ and $\mathbf{X}_2$ are IID samples, we have that,
\begin{align}
    &\mathbb{E}_{\mathbf{X}_1, \mathbf{X}_2 \sim \mathcal{D}(\mathbf{X})}[(f_{\theta}(\mathbf{X}_1) - f_{\theta}(\mathbf{X}_2))^2]\\
    &= \mathbb{E}_{\mathbf{X}_1, \mathbf{X}_2 \sim \mathcal{D}(\mathbf{X})}[( (f_{\theta}(\mathbf{X}_1) - \mathbf{\mu})  - (f_{\theta}(\mathbf{X}_2) - \mathbf{\mu}) )^2]\\
    &= \mathbb{E}_{\mathbf{X}_1 \sim \mathcal{D}(\mathbf{X})}[ (f_{\theta}(\mathbf{X}_1) - \mathbf{\mu})^2] +  \mathbb{E}_{\mathbf{X}_2 \sim \mathcal{D}(\mathbf{X})}[(f_{\theta}(\mathbf{X}_2) - \mathbf{\mu} )^2] \nonumber\\
    & - 2 \mathbb{E}_{\mathbf{X}_1,\mathbf{X}_2 \sim \mathcal{D}(\mathbf{X})}[ (f_{\theta}(\mathbf{X}_1) - \mathbf{\mu}) (f_{\theta}(\mathbf{X}_2) - \mathbf{\mu}) ]\\
    &= 2\mathbb{E}_{\mathbf{X} \sim \mathcal{D}(\mathbf{X})}[ (f_{\theta}(\mathbf{X}) - \mathbf{\mu})^2]\\
    &= 2\sigma^2
\end{align}
which proves yields the claim. $\square$
\end{proposition}

\end{document}